\documentclass[11pt]{article}

\usepackage[T1]{fontenc}
\usepackage[utf8]{inputenc}
\usepackage[margin=1in]{geometry}
\usepackage{lmodern}
\usepackage{microtype}
\usepackage{amsmath,amssymb,amsthm}
\usepackage{mathtools}
\usepackage{array}
\usepackage{tabularx}
\usepackage{booktabs}
\usepackage{graphicx}
\usepackage{float}
\usepackage{enumitem}
\usepackage{fancyvrb}
\usepackage[numbers,sort&compress]{natbib}
\usepackage[hidelinks]{hyperref}

\hypersetup{
  pdftitle={Auditing Representation-Space Fraud Detection: A Bayesian IRL Framework for Calibrated Uncertainty and Its Limits},
  pdfauthor={Shivam Swarup, Divya Prakash Shrivastava, Rakesh Thakur}
}
\newtheorem{proposition}{Proposition}
\newtheorem{corollary}{Corollary}

\newcommand{\cfg}[1]{\texttt{#1}}

\title{Auditing Representation-Space Fraud Detection:\\
A Bayesian IRL Framework for Calibrated Uncertainty and Its Limits}
\author{Shivam Swarup\\
JAIN University (Deemed-to-be)\\
Bengaluru, India\\
\texttt{shivam@jainuniversity.ac.in}
\and
Divya Prakash Shrivastava\\
Zayed University\\
Dubai, UAE\\
\texttt{Divya.shrivastava@zu.ac.ae}
\and
Rakesh Thakur\\
JAIN University (Deemed-to-be)\\
Bengaluru, India}
\date{}

\begin{document}
\maketitle

\begin{abstract}
Shilling detectors for recommender systems are almost always evaluated by one number, detection AUROC, with no account of when it can be trusted. We audit representation-space detection with VEIL (Variational Evaluation via Inverse-reinforcement Learning), a pairwise (Bradley--Terry) Bayesian reward-inference framework over frozen encoder embeddings, after the Bayesian auditing paradigm for LLM alignment~\cite{Bou2026}: (A)~a base posterior; (B)~a hierarchical ARD extension with a learned confidence; and (C)~a label-routed organic/attacker mixture used as an evidence-accumulation check. We prove (Prop.~\ref{prop:invariance}) that no detector built on a representation invariant to some transformation of user profiles can beat chance against an attack indistinguishable from organic profiles under that invariance. Detection is near-perfect against synthetic-structure attacks (AUROC 0.936--1.000) but collapses under donor-mimicry, which clones a real user's item set and splices in 1--3 target items: to chance on MovieLens-1M (0.48--0.51), partially on Amazon Musical Instruments, and least on CiaoDVD. Donor-mimicry is not undetectable: a non-learned baseline, each profile's maximum raw item-set Jaccard similarity to any other, reaches AUROC 0.92--0.97 (average precision 0.50--0.57) on all four donor-mimicry configurations, including ML-1M, where every embedding-based score is at chance. The signal is in the raw profiles; LightGCN's representation discards it, placing injected profiles closer to their true donor than its zero-propagation MF reduction does (distance 1.98 vs.\ 3.09). Severity varies by dataset, encoder, and detection method, with no one axis predicting the others. The limit lies in the representation, not the attack, and raw-similarity checks are a necessary complement to any learned detector.
\end{abstract}


\medskip

\section{Introduction}\label{sec:intro}

A shilling-detection paper is, almost without exception, evaluated by a single number: detection AUROC on whatever attacks the authors chose to inject. That number says nothing about \emph{when} it can be trusted: whether confidence tracks the actual difficulty of the attack, or whether a slightly more realistic attack would collapse it to chance unnoticed. Table~\ref{tab:ladder} shows exactly this in our own pipeline: every synthetic and GAN-generated attack we test -- classic random/average/bandwagon injection, AUSH, and AUSH with per-profile rating-style heterogenization -- is detected at AUROC 0.936--1.000 regardless of which of our three inference components is used, and a single change to the attack (cloning a real user's item-set and rating history and splicing in only 1--3 target items) drives detection to AUROC 0.48--0.51: exact chance, across all three components and an 8-seed sweep. A detector reporting only the first number would have shipped, unqualified, a system blind to the second attack.

This gap is not primarily a modeling problem to be closed with a better detector; it is structural. Proposition~\ref{prop:invariance} proves that any detector, however trained or complex, operating on a representation invariant to some transformation of user profiles cannot beat chance once an attack's profiles are indistinguishable under that invariance -- a property of the representation, not a claim that the attack itself is undetectable (\S\ref{sec:invariance}). This holds for every classifier in this paper, and for the widely-used label-free encoder we test (\S\ref{sec:mechanism}) it is not hypothetical: donor-mimicry attacks, constructed to approximately satisfy the proposition's hypothesis, drive detection from near-ceiling to exact chance. Nor is severity predictable from any single axis of comparison: \S\ref{sec:arch} shows it varies by dataset, encoder architecture, and detection method, with a ranking on one axis not transferring once a second also varies. Given a failure mode that is provable, empirically confirmed, and not bounded by checking any one dimension, the right response is not a detector claiming to have solved it but an auditing framework that reports calibrated confidence in its own detections and is checked against exactly these axes. VEIL, our BIRL-style auditing framework, is built for that job. Its three-stage structure -- quantify ambiguity in a posterior, audit that posterior's trustworthiness via calibrated epistemic uncertainty, validate the audit against evidence accumulation -- follows the shift of Bou et al.'s Alignment Auditor for verifying LLM objectives~\cite{Bou2026}, which we adapt to shilling detection: ``is this user manipulating the system'' becomes a Bayesian reward to infer, not a label to predict.

Concretely, we fit a static pairwise Bradley--Terry reward model by variational inference over frozen recommender embeddings (\S\ref{sec:veil}) in three components: (A)~a base posterior with fixed confidence; (B)~a hierarchical extension with per-dimension relevance priors and a learned confidence $\alpha$ as the calibrated-uncertainty audit signal; and (C)~a two-branch organic/attacker mixture whose attacker-branch posterior we watch contract as evidence accumulates. Component~(C) routes profiles by their labels, so it is an auditing experiment, not an unsupervised detector (a label-free attempt yields only a weakly attack-related partition; Appendix~\ref{app:unsup}). We run all three on a supervised GNN (Sup-GNN, trained on the attack labels: a circular positive control) and a label-free LightGCN (the honest test).

Running VEIL end to end surfaces three findings, plus a validation check: Component~C's attacker-branch posterior contracts monotonically as injected-profile evidence accumulates, decoupled from point-estimate accuracy (\S\ref{sec:contraction}) -- expected Bayesian behavior, confirming the machinery works rather than a finding in its own right. First, and unexpected: Component~B's learned $\alpha$ varies across the attack-difficulty ladder but not as a clean ``confident when easy, unsure when hard'' dial -- lowest on the easiest configurations, peaking on the hardest attack still well above chance, and intermediate once detection has collapsed (\S\ref{sec:alpha}). Second, the donor-mimicry collapse has a specific, traceable mechanism on our label-free encoder: LightGCN's adjacency construction and BPR loss never read a rating value, only item identity (\S\ref{sec:mechanism}), so no realism added to an attack's rating \emph{values} could have mattered. This is not a new vulnerability class~\cite{Yu2006,Yu2008,Fan2021,Zhang2021}; our contribution is a mechanistic, calibrated account of where and why that limit bites a BIRL-style auditing framework, not its discovery. Third, the collapse is not a property of ML-1M, of LightGCN, or of VEIL's detection method alone: it is severe on ML-1M, partial on Musical Instruments, and largely absent on CiaoDVD (\S\ref{sec:datasets}); the matrix-factorization reduction of LightGCN is more robust than LightGCN on ML-1M's hardest point but less robust on both CiaoDVD points; and VEIL beats a classical RDMA/WDMA baseline on CiaoDVD but is comparable to it on ML-1M, where both collapse (\S\ref{sec:arch}). No one axis predicts the others, which is the empirical counterpart to Proposition~\ref{prop:invariance}: a detectability limit whose severity cannot be assumed to transfer without checking.

\section{Related Work}\label{sec:related}

\paragraph{Shilling attacks and detection.} Profile-injection attacks on collaborative filtering were first demonstrated by \citet{LamRiedl2004}; Random, Average, and Bandwagon were formalized, together with an analysis of algorithm robustness against them, by \citet{Mobasher2007}. Classical detectors classify users from hand-crafted rating-statistics features such as RDMA and WDMA~\cite{Chirita2005}, which we use as a baseline (\S\ref{sec:arch}). Attack generation has since moved to GAN-based profile synthesis (AUSH;~\cite{Lin2020}), GAN-plus-graph-convolution attacks~\cite{Wu2021}, and diffusion-based generators~\cite{Qiao2025}. Recent detectors include multi-view cross-attention~\cite{Zhai2026}, credibility-weighted matrix factorization~\cite{Zhao2025}, federated~\cite{Jiang2020}, graph-convolutional group~\cite{Wang2022}, collusive-behavior group~\cite{Xu2023}, and temporal/relational~\cite{Zhang2025} methods. Our concern is complementary to any one of these: not a new detector, but when a detector's reported accuracy can be trusted.

\paragraph{Bayesian inverse reinforcement learning.} Component~B's learned confidence and per-dimension ARD structure follow the hierarchical-prior direction of \citet{ChoiKim2015}, reduced here to an empirical-Bayes/ARD treatment with no MDP, Q-function, or policy solving (\S\ref{sec:prelim}). \citet{ImaniGhoreishi2021} address scalability of Bayesian IRL, and \citet{Hossain2022} quantify how much behavioral data a Bayesian IRL model needs, the sample-size analogue of the posterior-ambiguity question Component~A asks.

\paragraph{Sybil defense and profile-copying attacks.} The donor-mimicry result (\S\ref{sec:results}) confirms a known limit class rather than introducing a new attack. SybilGuard and SybilLimit~\cite{Yu2006,Yu2008} are the foundational social-graph Sybil defenses. CopyAttack~\cite{Fan2021} is the closest analogue to donor-mimicry in the recommender-attack literature: it copies real user profiles across domains to evade detectors that expect synthetic structure, the mechanism our donor-mimicry attack uses within a single domain (\S\ref{sec:mechanism}). RecUP~\cite{Zhang2021} is a GAN-based poisoning attack designed so injected profiles' summary statistics stay indistinguishable from organic ones, the same evasion goal donor-mimicry pursues by copying real profiles outright rather than approximating their statistics.

\paragraph{Bayesian uncertainty in recommender systems.} To our knowledge, calibrated posterior inference has not been applied to shilling detection or attacker-profile identification; the closest work, \citet{Chen2021}, learns reward functions adversarially but does not quantify uncertainty. We flag this as a gap in our literature search, not evidence that no such work exists.

\paragraph{Auditing frameworks.} The Alignment Auditor~\cite{Bou2026} verifies LLM objectives by inferring a Bayesian posterior over them and auditing that posterior's uncertainty. Our three-stage structure (quantify a posterior, audit its calibrated uncertainty, validate against evidence accumulation) follows that template (\S\ref{sec:veil}); what we add is the setting, the invariance result of \S\ref{sec:invariance}, and the cross-axis empirical checks, not the template itself.

\section{Preliminaries / Problem Setup}\label{sec:prelim}

We adopt a \textbf{static pairwise Bradley--Terry} reformulation over frozen embeddings rather than a sequential MDP-IRL formulation, because MovieLens-1M~\cite{HarperKonstan2015} interaction logs are not agent \emph{trajectories} in any sense that supports a clean identifiability argument: there is no well-defined action sequence, state transition, or ``demonstrator policy'' to invert. Casting the problem as \emph{static preference inference} over paired profile embeddings avoids that objection, at the declared cost (\S\ref{sec:limitations}) of not modeling how ratings accrue over time.

Formally: for each profile $x$ (organic or injected) we have a frozen embedding $\phi(x)\in\mathbb{R}^{64}$ (LightGCN) or $\mathbb{R}^{64\text{ or }128}$ (Sup-GNN raw HGT / post-fusion). We form pairs $(x^+,x^-)$ of matched organic/injected profiles (random or $k$-NN matching) and posit a Bradley--Terry preference likelihood
\begin{equation}
P(x^+\succ x^-\mid\theta)=\sigma\Big(\alpha\cdot\theta^\top\big(\phi(x^+)-\phi(x^-)\big)\Big)
\end{equation}
over a linear reward $r_\theta(x)=\theta^\top\phi(x)$: a reward-inference problem with no MDP, Q-function, or policy solving. Without an MDP, this reduces to Bayesian logistic regression on embedding differences; we retain the IRL terminology because the structure follows the Bayesian reward-inference auditing template of \citet{Bou2026}, not because we solve a sequential IRL problem.

\section{VEIL -- Components A/B/C as Stages 1--3}\label{sec:veil}

\paragraph{Stage 0 (shared).} Extract frozen $\phi$ for every profile from one of two encoders: a Sup-GNN detector (trained with focal loss \emph{on the injected labels}, so $\phi$ already encodes organic-vs-injected, making it a circular positive-control encoder), or a LightGCN recommender, trained purely with BPR ranking loss over the poisoned bipartite graph and blind to the organic/injected label by construction (\S\ref{sec:mechanism} gives the mechanism). Standardize with train-pool mean/std and cache.

\paragraph{Stage 1 -- Component A: quantify ambiguity.} Fixed-confidence pairwise BIRL. Prior $\theta\sim\mathcal{N}(0,\sigma_0^2 I)$, mean-field posterior $q(\theta)=\mathcal{N}(\mu,\operatorname{diag}(\sigma^2))$, fit by the reparameterization-trick ELBO
\begin{equation}
\mathbb{E}_q\Big[\textstyle\sum\log\sigma\big(\alpha\,\theta^\top\Delta\phi\big)\Big]-\mathrm{KL}\big(q\,\|\,\mathcal{N}(0,\sigma_0^2 I)\big)
\end{equation}
with $\alpha\equiv1$ fixed and minibatch Adam (settings in Appendix~\ref{app:hyper}). The posterior's marginal covariance log-determinant and $\mu,\sigma$ spread (per config in Appendix~\ref{app:compA}) quantify the ambiguity: how tightly the evidence pins down the reward direction.

\paragraph{Stage 2 -- Component B: audit trustworthiness via calibrated epistemic uncertainty.} Extends A with a learned confidence $\alpha\sim\mathrm{Gamma}(\tau_1,\tau_2)$ and per-dimension ARD precisions $\lambda_d\sim\mathrm{Gamma}(\gamma_1,\gamma_2)$ shaping $\mu_d\sim\mathcal{N}\big(0,(\beta\lambda_d)^{-1}\big)$, jointly optimized on the extended ELBO and warm-started from A's posterior (hyperparameters: Appendix~\ref{app:hyper}). The learned $\alpha$ is the audit signal: how much of the pairwise log-odds the model will stake its confidence on, a variational posterior quantity rather than a fixed hyperparameter. \S\ref{sec:alpha} reports what it tracks across the ladder: a real but non-monotonic pattern.

\paragraph{Stage 2/3 -- Component C: posterior contraction as evidence accumulates.} Ground-truth labels route profiles into two independent B-style branches, organic-anchored and attacker-anchored, each fit exactly as B. Held-out detection compares branch reward projections $r_{\mathrm{att}}(x)-r_{\mathrm{org}}(x)$, calibrated to $P(\text{injected})$ by a 1-D logistic fit on the train split. Because routing uses the labels, Component~C is an auditing experiment, not an unsupervised detector; an inferred responsibility $q(z_i)$ converges stably but to a partition only weakly tied to the attackers (Appendix~\ref{app:unsup}). \S\ref{sec:contraction} checks that the attacker branch's posterior (looser than the organic branch's, as the minority class) contracts monotonically with more injected-profile evidence, as correct Bayesian behavior requires. Algorithm boxes: Appendix~\ref{app:algo}.

\section{Experimental Setup}\label{sec:setup}

\begin{itemize}[leftmargin=*,itemsep=2pt]
\item \textbf{Dataset.} MovieLens-1M~\cite{HarperKonstan2015}, 5-core filtered (999,611 interactions after filtering, from 1,000,209 raw), subsampled to 30\% of users before injection.
\item \textbf{Encoders.} Sup-GNN -- supervised, trained on injected labels via focal loss; a positive control, expected to be circular. LightGCN -- label-free, BPR ranking loss only; the honest baseline. Both were trained from scratch: Sup-GNN 20 epochs, LightGCN 30 BPR epochs, dim 64, 3 layers.
\item \textbf{Attack ladder (5 run configs).} Five configs of increasing difficulty, sharing Sup-GNN/\allowbreak LightGCN encoders: classic random\,+\,average\,+\,bandwagon injection; AUSH (GAN-generated profiles); AUSH wrapped in a heterogenization attack (organic-matched profile/filler/rating-style variation); the same plus donor-mimicry (each fake profile copies a real organic user's item-set and ratings, splicing in 1--3 target items); and an 8-seed sweep repeating the donor-mimicry config to quantify variance at the collapse point. Results in Table~\ref{tab:ladder}; per-config parameters in Appendix~\ref{app:hyper}.
\item \textbf{Compute.} CPU-only, Intel i3-1115G4 (2 cores / 4 threads), 8\,GB RAM, no GPU. Wall-clock per stage: Appendix~\ref{app:wallclock}.
\item \textbf{Metrics.} Pairwise accuracy, Brier, ECE on held-out preference pairs; AUROC, AP, accuracy, F1/precision/recall on held-out organic-vs-injected detection; ELBO and posterior log-determinant as ambiguity/contraction diagnostics.
\end{itemize}

\section{Results and Discussion}\label{sec:results}

\subsection{Full attack-difficulty ladder}\label{sec:ladder}

All values are held-out detection metrics (Components A/B) or combined detection metrics (Component~C), rounded to 3 significant figures for AUROC/AP and 2 decimals for $\alpha$; the supplementary materials give full-precision values.

\paragraph{Structural/synthetic attacks (rows 1--4): AUROC 0.936--1.000.} The floor across A/B/C on this segment is $0.936$ (Component~C, \cfg{lightgcn\_aush\_het}), driven by the more sophisticated heterogenized attack -- not a uniform ``regardless of sophistication'' ceiling.

\paragraph{Donor-mimicry (row 5): collapses to chance.} At seed 42, Component~A AUROC is $0.500$ and AP $0.051$ (base rate $\approx 27/571\approx4.7\%$ positive class, so AP near the base rate is exactly the chance-level signature). Across the full 8-seed sweep, AUROC (mean $\pm$ population std) is $0.494\pm0.061$ (A), $0.488\pm0.062$ (B), and $0.481\pm0.058$ (C).

\begin{table}[t]
\centering
\footnotesize
\setlength{\tabcolsep}{3.5pt}
\begin{tabular}{lccccccc}
\toprule
config & attack & A AUROC & A AP & B AUROC & B $\alpha$ & C AUROC & C $\alpha_{\mathrm{org}}/\alpha_{\mathrm{att}}$\\
\midrule
\cfg{supgnn\_base} & random+avg+bw & 0.998 & 0.961 & 0.998 & 3.22 & 0.998 & 3.24 / 3.00\\
\cfg{lightgcn\_base} & random+avg+bw & 0.999 & 0.995 & 1.000 & 0.42 & 1.000 & 0.47 / 0.30\\
\cfg{lightgcn\_aush} & AUSH & 1.000 & 1.000 & 1.000 & 0.40 & 1.000 & 0.69 / 0.31\\
\cfg{lightgcn\_aush\_het} & AUSH+Het. & 0.942 & 0.423 & 0.944 & 0.70 & 0.936 & 0.83 / 0.49\\
\cfg{lightgcn\_aush\_het\_donor} & +Donor-mimicry & \textbf{0.500} & 0.051 & 0.513 & 0.54 & 0.507 & 0.54 / 0.39\\
\bottomrule
\end{tabular}
\caption{Attack-difficulty ladder, LightGCN honest baseline + Sup-GNN circular control; all columns are seed-42 runs (8-seed $\alpha$ means: \S\ref{sec:alpha}; donor-mimicry 8-seed AUROCs: text above).}
\label{tab:ladder}
\end{table}

\subsection{Calibration tracks (and sharpens) the same collapse}\label{sec:calibration}

AUROC alone cannot say whether a detector's confidence is trustworthy where it fails. We run an 8-seed sweep on all ten configs (five ladder points, five cross-dataset points of \S\ref{sec:datasets}) and report profile-level Brier, NLL, and ECE via the same 1-D logistic calibration Component~C uses (\S\ref{sec:veil}), with Student-$t$ 95\% CIs. Table~\ref{tab:calib} shows a subset for Component~A; all configs and reliability bins: Appendix~\ref{app:calib}.

\begin{table}[t]
\centering
\small
\begin{tabular}{lcc}
\toprule
config & AUROC (95\% CI) & ECE (95\% CI)\\
\midrule
\cfg{lightgcn\_base} (easy) & 1.000 [0.999, 1.000] & 0.006 [0.004, 0.008]\\
\cfg{lightgcn\_aush\_het} & 0.945 [0.932, 0.958] & 0.090 [0.082, 0.099]\\
CiaoDVD (easy) & 0.877 [0.866, 0.888] & 0.177 [0.171, 0.184]\\
Musical Instruments (hard) & 0.805 [0.797, 0.814] & 0.263 [0.259, 0.268]\\
\cfg{lightgcn\_aush\_het\_donor} (hard) & 0.494 [0.440, 0.549] & 0.374 [0.360, 0.388]\\
\bottomrule
\end{tabular}
\caption{Component~A, profile-level AUROC and ECE, 8-seed mean [95\% CI], five of ten swept configs (the rest: Appendix~\ref{app:calib}).}
\label{tab:calib}
\end{table}

\begin{figure}[H]
\centering
\includegraphics[width=\textwidth]{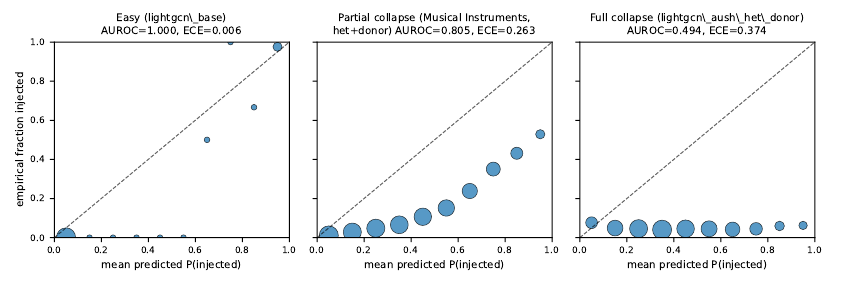}
\caption{Pooled (8-seed) reliability diagrams, Component~A: easy, partial collapse (Musical Instruments, matched hard config), and full collapse (ML-1M full-ladder donor-mimicry). Point size $\propto$ bin count. Dashed line: perfect calibration.}
\label{fig:reliability}
\end{figure}

ECE moves in lockstep with AUROC and amplifies the story: every swept config with AUROC $\ge0.94$ has ECE $\le0.09$, while every config below $0.88$ has ECE $0.18$--$0.37$ -- the detector is not merely wrong at the hard end, it is \emph{confidently} wrong. The reliability diagrams (Figure~\ref{fig:reliability}, Appendix~\ref{app:calib}) show this: the easy config sits near the diagonal's extremes, the partial-collapse config is uniformly \emph{below} the diagonal (overconfident), and the full-collapse config is flat near the base rate -- the calibrated probability carries no information, as Proposition~\ref{prop:invariance} predicts.

\subsection{Component B's learned confidence \texorpdfstring{$\alpha$}{alpha} -- non-monotonic, flagged}\label{sec:alpha}

Within the LightGCN honest-baseline ladder (8-seed mean $\pm$ SE), $\alpha$ is \textbf{lowest on the two easiest configs} (\cfg{lightgcn\_base} $0.386\pm0.006$, \cfg{lightgcn\_aush} $0.329\pm0.021$) and \textbf{highest on the hardest \emph{non-collapsed} config} ($0.688\pm0.022$ on \cfg{lightgcn\_aush\_het}), then drops back to an intermediate $0.551\pm0.030$ where detection has fully collapsed to chance. This is an inverted-U over the four ladder points, peaking at the hard-but-still-separable attack rather than the collapsed one; the peak's 95\% CI ($[0.637, 0.739]$) overlaps neither the collapsed point's ($[0.480, 0.623]$) nor the easy points', though the ordering of the two easy points is not resolved (their CIs overlap slightly). A plausible, post-hoc reading: $\alpha$ tracks how much pairwise log-odds signal remains for the model to stake weight on, which is not monotonic in AUROC-defined difficulty. The spreads cover inference seeds over one trained encoder per config only (\S\ref{sec:limitations}).

\subsection{Component C -- posterior contraction as a validation check}\label{sec:contraction}

A sanity check, not a novel finding: posterior contraction under more evidence is what a correctly-implemented Bayesian model should do. In an evidence-count ablation on the base LightGCN config (full table: Appendix~\ref{app:compC}, Table~\ref{tab:ablation}), from $n_{\text{injected}}=20$ to $191$, $\log|\Sigma|$ falls monotonically ($78.49\to15.95$), mean $\sigma$ falls ($1.858\to1.139$), and $\alpha$ rises ($0.183\to0.297$), while branch-detection AUROC stays saturated ($0.999\to1.000$): the contraction reflects how confident the posterior is entitled to be, decoupled from whether the point estimate already detects well.

\paragraph{Does this require ground-truth routing?} We tested replacing label-based routing with a fully label-free stochastic variational-EM procedure (Appendix~\ref{app:unsup}). The runs converge to one partition with roles swapped (label switching, unresolvable by the ELBO); a label-free rule (smaller cluster = attacker) gives a stable post-hoc AUROC of about $0.61$, but the partition agrees only weakly with the true labels (ARI $0.05$): a partial capability, not a drop-in replacement.

\subsection{Why LightGCN goes blind under donor-mimicry: a mechanistic account}\label{sec:mechanism}

Two functions in \texttt{src/models/lightgcn.py} explain why: the adjacency builder collapses every (user, item) pair to one unweighted edge regardless of rating value, and the BPR loss takes only item IDs, never a rating tensor, so rating \emph{value} is structurally never read (line-level citations in Appendix~\ref{app:hyper}). The LightGCN embedding is therefore a function purely of \emph{which items} a profile rated (and, through propagation, which other users share them), never \emph{how} they rated them. Donor-mimicry exploits exactly this: it copies a real organic user's full item-set verbatim and splices in only 1--3 target items, so the profile is, on the only axis LightGCN can see, nearly indistinguishable from an organic one. The rating-value levers (per-profile style, filler noise, target-rating jitter) that made \cfg{lightgcn\_aush\_het} harder than \cfg{lightgcn\_aush} do \textbf{not} move LightGCN separability beyond what item-set structure gives; donor-mimicry is the first lever that also perturbs the item-set axis, which is why it -- and only it -- collapses detection.

\paragraph{This is not a novel discovery.} That representation-space Sybil/shilling detectors are vulnerable to attacks that copy real user behavior is well established~\cite{Yu2006,Yu2008,Fan2021,Zhang2021}; our contribution is showing \emph{precisely}, down to the two functions above, why this encoder is blind to it.

\paragraph{Direct measurement, not just code inspection.} We also tested the propagation-depth account directly. For each donor-mimicry-injected profile in the ML-1M hard-ladder point we recovered its true donor (not stored by the attack; reconstructed by item-set Jaccard overlap, median $0.96$, with 81 of 91 profiles confidently matched) and compared injected-to-donor embedding distance under LightGCN and under its $n_{\text{layers}}=0$ reduction (matrix factorization; \S\ref{sec:arch}) on the identical poisoned dataset, with no retraining. LightGCN places injected profiles reliably closer to their true donor than MF does (mean Euclidean distance $1.98$ vs.\ $3.09$; paired difference $1.11$, $n=81$). SEMs, cosine similarities, and a random-organic-user control, which rules out a general embedding-scale difference between the two fits, are in Appendix~\ref{app:mf}. This is consistent with, though on a single config not sufficient alone to establish causally, the propagation-depth account: deeper message passing appears to smooth an injected profile toward its donor, erasing the residual signal Proposition~\ref{prop:invariance} identifies as the only thing donor-mimicry leaves behind. The same architecture signature -- LightGCN, the only propagating encoder among four tested, scoring lowest -- reappears on a second dataset, Musical Instruments, once two non-propagating sequential encoders are added alongside MF (\S\ref{sec:arch}). The donor closeness visible here in an embedding is visible without one too: raw item-set Jaccard detects these profiles directly, AUROC 0.92--0.97 where LightGCN collapses (Appendix~\ref{app:labelfree}) -- a limit of what LightGCN discards, not of the attack (\S\ref{sec:invariance}).

\subsection{Formalizing the invariance argument}\label{sec:invariance}

\S\ref{sec:mechanism} shows, at the level of two specific functions, that LightGCN's representation is blind to rating values. We now state the general property this instantiates, so the mechanism is legible as a property of \emph{any} representation with the same invariance, not an artifact of one implementation.

Let $X$ be the space of user profiles and $\pi:X\to S$ a map onto the ``structural signature'' a representation actually consumes -- for LightGCN, $\pi(x)$ is the deduplicated (item, co-occurrence) structure of $x$'s ratings, discarding rating \emph{values} entirely (exactly what \texttt{build\_norm\_adj}/\texttt{bpr\_loss} do; \S\ref{sec:mechanism}). Say a representation $\phi:X\to\mathbb{R}^d$ is \emph{$\pi$-invariant} if it factors as $\phi=\bar\phi\circ\pi$ for some $\bar\phi:S\to\mathbb{R}^d$, i.e.\ $\phi(x)$ depends on $x$ only through $\pi(x)$.

\begin{proposition}[Representation-invariance undetectability]\label{prop:invariance}
Let $\phi$ be $\pi$-invariant, $Y\in\{0,1\}$ the organic/attacker label, and $D:\mathbb{R}^d\to\mathbb{R}$ any measurable detector score (this includes every detector in this paper -- Components A/B/C's reward projections -- and any other function of $\phi(X)$ whatsoever, linear or not, Bayesian or not). If an attack produces profiles $X$ such that $\pi(X)\mid Y{=}1$ and $\pi(X)\mid Y{=}0$ are identically distributed on the attack's support, then $D(\phi(X))$ has identical distribution under $Y{=}1$ and $Y{=}0$ on that support, and $D$ achieves AUROC $=0.5$ exactly, regardless of $D$'s complexity or how it was fit.
\end{proposition}

\begin{proof}[Proof sketch]
$\phi=\bar\phi\circ\pi$ is deterministic, so pushing the (by hypothesis, equal) conditional distributions of $\pi(X)$ under $Y{=}0$ and $Y{=}1$ through $\bar\phi$ gives $\phi(X)\mid Y{=}1\overset{d}{=}\phi(X)\mid Y{=}0$, i.e.\ $\phi(X)\perp Y$ on the attack's support, so $I(Y;\phi(X))=0$. By the data-processing inequality, for any (possibly stochastic) $D$, $I(Y;D(\phi(X)))\le I(Y;\phi(X))=0$, and mutual information is non-negative, so $I(Y;D(\phi(X)))=0$: the score $D(\phi(X))$ is statistically independent of $Y$, hence identically distributed under both labels. AUROC $=P(\mathrm{score}_1>\mathrm{score}_0)+\tfrac12P(\mathrm{tie})$ for two independent draws from the \emph{same} distribution is exactly $0.5$ by the symmetry of that draw.
\end{proof}

\begin{corollary}\label{cor:agnostic}
The bound in Proposition~\ref{prop:invariance} is architecture-agnostic: it holds for Components A/B/C, for the auditing machinery built around them, and equally for any other classifier trained on $\phi(X)$ -- calibration, uncertainty quantification, or ensembling cannot recover signal the representation never carried.
\end{corollary}

\paragraph{What this does and does not claim.} The hypothesis -- that an attack achieves \emph{exactly} matching structural distributions -- is an existence claim about attacks, not a property of every attack. Donor-mimicry approximately satisfies it: it copies a real user's full item-set verbatim and splices in only 1--3 target items (\S\ref{sec:mechanism}), so $\pi(X)$ matches an organic profile's on every coordinate except those few items. The proposition therefore predicts AUROC $\to0.5$ as the splice fraction vanishes, not that every run hits exactly $0.5$: the shared target items across attacker profiles are a small residual violation, consistent with results that are \emph{close to} but not always at chance (0.48--0.54 across configs and datasets, \S\ref{sec:ladder}, \S\ref{sec:datasets}; exactly $0.500$ only at seed 42 on the original ladder). We report the proposition as an idealized limiting case that the empirical mechanism approaches, not one donor-mimicry achieves exactly.

\paragraph{The bound is about the representation, not the attack.} Proposition~\ref{prop:invariance} bounds detectors that are functions of a $\pi$-invariant $\phi$, not a detector on the raw profile, which need not share $\phi$'s invariance. Raw item-set Jaccard similarity -- no embedding -- detects donor-mimicry at AUROC 0.92--0.97 on every config where LightGCN and every embedding-based detector, VEIL included, sit at or near chance (Appendix~\ref{app:labelfree}): the profiles are near-duplicates of a specific organic user, trivially visible in the raw data. LightGCN's \emph{particular} invariance (item-ID structure smoothed by propagation) discards a signal a coarser, embedding-free view retains. ``Detectability limit'' here is a property of the representations we test, not a claim that donor-mimicry defeats detection outright.

\subsection{Does the donor-mimicry collapse generalize across datasets?}\label{sec:datasets}

A single-dataset ladder cannot say whether the collapse reflects donor-mimicry in general or ML-1M's embedding geometry. We run the same LightGCN + Components A/B/C pipeline on two further datasets -- Amazon Musical Instruments~\cite{Hou2024} and CiaoDVD, a trust-aware DVD-rating dataset~\cite{Guo2014}, each at its own post-5-core-filter scale (sizes in Appendix~\ref{app:hyper}) -- holding the attack recipe fixed: classic Random+Average+Bandwagon for the easy point, classic+heterogenize+donor-mimicry for the hard point, deliberately without AUSH.

The original ML-1M donor-mimicry point (\S\ref{sec:ladder}; AUSH+heterogenize+donor-mimicry, AUROC 0.48--0.51) is not comparable to the other datasets' hard points, which never used AUSH, so we also ran ML-1M under the matched recipe (\textbf{matched config}: no AUSH), giving AUROC 0.530--0.536 (8-seed mean; Table~\ref{tab:datasets}), a small, seed-consistent margin above the ``full ladder'' number.

The result does not replicate uniformly. On \textbf{ML-1M} (matched), detection collapses from near-ceiling to chance-adjacent (0.999--1.000 $\to$ 0.530--0.536). On \textbf{Musical Instruments} the collapse is partial (0.991--0.992 $\to$ 0.805--0.808): a real drop ($\approx0.18$), well above chance. On \textbf{CiaoDVD} it is largely absent (0.877--0.879 $\to$ 0.846--0.848), because the classic-attack baseline never reached near-ceiling, leaving little headroom. Representation-space detectability limits under donor-mimicry are therefore dataset-dependent, not universal, and cannot be assumed to transfer without testing. What drives the divergence is open: neither catalog size nor rating density explains it (\S\ref{sec:limitations}).

\begin{table}[t]
\centering
\small
\begin{tabular}{llccc}
\toprule
Dataset & Point & A AUROC & B AUROC & C AUROC\\
\midrule
ML-1M & easy & 0.999 & 1.000 & 1.000\\
ML-1M & hard (matched, no AUSH)$^\dagger$ & $0.533\pm0.028$ & $0.536\pm0.037$ & $0.530\pm0.030$\\
Musical Instruments & easy & 0.991 & 0.992 & 0.991\\
Musical Instruments & hard$^\dagger$ & $0.805\pm0.009$ & $0.808\pm0.009$ & $0.808\pm0.010$\\
CiaoDVD & easy$^\dagger$ & $0.877\pm0.012$ & $0.878\pm0.010$ & $0.879\pm0.012$\\
CiaoDVD & hard$^\dagger$ & $0.846\pm0.008$ & $0.848\pm0.007$ & $0.846\pm0.010$\\
\bottomrule
\end{tabular}
\caption{Three-dataset generalization check: easy (classic attacks) vs.\ hard (classic+heterogenize+donor-mimicry, no AUSH) AUROC, LightGCN encoder. $^\dagger$8-seed mean$\pm$std; unmarked rows are single runs (\S\ref{sec:limitations}). ML-1M's hard point is the matched-recipe config, distinct from the full-ladder point in Table~\ref{tab:ladder}.}
\label{tab:datasets}
\end{table}

\subsection{Does the collapse depend on architecture or detection method too?}\label{sec:arch}

\S\ref{sec:datasets} varies the dataset while holding architecture (LightGCN) and detection method (VEIL's Components A/B/C) fixed. We now vary those two axes independently against the same donor-mimicry attack, to test whether either produces a stable ranking on its own.

\paragraph{Architecture.} We reran the full ladder with LightGCN's propagation depth set to $n_{\text{layers}}=0$, an exact reduction to classical BPR matrix factorization~\cite{Rendle2009} (confirmed by training-loop inspection; not a separately implemented architecture). Easy and AUSH points are indistinguishable between the two (both near-ceiling). On ML-1M's donor-mimicry point MF is \emph{substantially more robust} than LightGCN (AUROC 0.632 vs.\ 0.481--0.494). On both CiaoDVD points the ranking reverses: LightGCN is \emph{more} robust (classic: 0.877--0.879 vs.\ 0.814; donor-mimicry: 0.846--0.848 vs.\ 0.764), plausibly because MF underfits CiaoDVD's much sparser per-user signal ($\approx15$ vs.\ $\approx165$ ratings/user). Full 6-config table: Appendix~\ref{app:mf}.

\paragraph{Sequential encoders.} A sequential encoder might resist donor-mimicry, since the attack clones a donor's item set but not necessarily its temporal order. We tested SASRec (causal self-attention, embedding dim 64, matching LightGCN/MF) and an order-agnostic mean-pooled control (same architecture, causal mask and position removed) on Musical Instruments; both are label-free with no free per-user parameter vector. The donor-mimicry generator used elsewhere in this paper does not preserve temporal order (items copied in arbitrary order, targets always appended last, synthetic regular-cadence timestamps) -- an artifact a sequential encoder can exploit. A 3-seed check confirms it: SASRec reaches AUROC 0.9997--0.9999 (SD $\approx0.0001$) on the original generator, while mean-pool on the identical profiles scores far lower (0.840--0.857) -- positional, not a genuine signal. We built an order-preserving variant (each target item takes the position/timestamp of the donor item it displaced; verified bit-identical to the original on every other attack/config). All results below use this corrected generator.

\begin{table}[t]
\centering
\small
\begin{tabular}{lccc}
\toprule
Architecture & A AUROC & B AUROC & C AUROC\\
\midrule
LightGCN & $0.800\pm0.010$ & $0.802\pm0.010$ & $0.803\pm0.011$\\
mean-pool & $0.874\pm0.006$ & $0.881\pm0.007$ & $0.885\pm0.007$\\
SASRec & $0.901\pm0.002$ & $0.901\pm0.002$ & $0.902\pm0.003$\\
MF & $0.940\pm0.002$ & $0.941\pm0.003$ & $0.941\pm0.003$\\
\bottomrule
\end{tabular}
\caption{Four-architecture comparison, Musical Instruments, order-preserving donor-mimicry, 8-seed mean$\pm$std. Ranking is MF $>$ SASRec $>$ mean-pool $>$ LightGCN on every component; MF's 95\% CI does not overlap SASRec's on any component (e.g.\ Component~C: $[0.938, 0.943]$ vs.\ $[0.900, 0.904]$).}
\label{tab:arch}
\end{table}

LightGCN -- the only architecture here that performs graph propagation -- is the clear outlier at the bottom, reproducing the ML-1M donor-distance signature on a second dataset; the secondary MF/SASRec/mean-pool ordering is measured but not mechanistically explained (Appendix~\ref{app:mf}).

\paragraph{Detection method.} We compared VEIL against a classical baseline (A2: per-user RDMA/\allowbreak WDMA/\allowbreak LengthVar rating-value statistics~\cite{Chirita2005} + logistic regression, same poisoned dataset and split). On ML-1M both collapse under donor-mimicry (A2: $0.982\to0.584$; VEIL: $1.00\to0.53$); on CiaoDVD A2 collapses \emph{more} (A2: $0.840\to0.617$; VEIL: $0.878\to0.847$). Full table: Appendix~\ref{app:a2}.

\paragraph{One unifying observation, not three.} Collapse severity has a real, measured ranking by dataset (ML-1M worst, Musical Instruments partial, CiaoDVD least; \S\ref{sec:datasets}), but that ranking does not transfer once a second axis varies: MF beats LightGCN on ML-1M's donor-mimicry point but loses to it on both CiaoDVD points, and VEIL beats the classical baseline on CiaoDVD but only matches it on ML-1M. Dataset, architecture, and detection method interact, so a result from varying one is not safe to generalize along the others without checking. See Appendix~\ref{app:labelfree}.

\section{Limitations}\label{sec:limitations}

\begin{itemize}[leftmargin=*,itemsep=2pt]
\item \textbf{Three datasets, partial ladder coverage, unresolved divergence.} \S\ref{sec:datasets} covers ML-1M, Amazon Musical Instruments, and CiaoDVD, but only 2 of the 5 ladder points (classic-easy and donor-mimicry-hard) were run on the two new datasets, and Table~\ref{tab:datasets}'s ML-1M-easy and Musical-Instruments-easy AUROCs are single runs (their 8-seed sweeps, used in \S\ref{sec:calibration}, agree to within 0.001). Whether the pattern holds on the missing points or on other domains is untested. What drives the divergence is unresolved: catalog size does not track collapse severity (CiaoDVD: 2,069 items, least collapse; ML-1M: 3,416, most; Musical Instruments: 27,637, between), and rating density (ML-1M $\approx165$, CiaoDVD $\approx15.6$, Musical Instruments $\approx8.9$ ratings/user) does not either. CiaoDVD's resistance is largely an artifact of weak baseline separability.
\item \textbf{No live or human study.} All attacks and ``organic'' profiles come from static, already-collected datasets.
\item \textbf{Static pairwise BIRL, not sequential MDP-IRL.} \S\ref{sec:prelim}: the framework cannot model how a profile accrues ratings over time.
\item \textbf{Encoder and baseline coverage remains narrow.} Sup-GNN and LightGCN are GNN-family encoders on the same bipartite graph, and the collapse is mechanistically explained for LightGCN specifically (\S\ref{sec:mechanism}); \S\ref{sec:arch} adds matrix factorization and finds a dataset-dependent, not uniform, relationship to LightGCN. We did not pursue modern learned shilling-detector baselines (our only baseline is the classical RDMA/WDMA classifier). \S\ref{sec:arch} adds SASRec and an order-agnostic mean-pooled control on Musical Instruments' donor-mimicry point, at one embedding dimension with early-stopped training and no hyperparameter sweep; this project's CPU-only setup (\S\ref{sec:setup}) still limited us to that one dataset/config, not ML-1M/CiaoDVD, and our claims are scoped accordingly.
\item \textbf{Component C's routing is ground-truth, not inferred.} It is an auditing experiment, not a deployable detector; our label-free attempt (Appendix~\ref{app:unsup}, one configuration) found only a weakly attack-related partition (ARI 0.05).
\item \textbf{Component B $\alpha$ spread covers inference seeds only.} The 8-seed spreads (\S\ref{sec:alpha}) vary pair sampling, split, and VI initialization over one trained encoder per config; encoder-retraining variance is not measured, and the inverted-U reading is post-hoc.
\end{itemize}

\section{Conclusion}\label{sec:conclusion}

We set out from a structural claim rather than a benchmark score: representation-space fraud detection built on an invariant embedding has a provable blind spot (Proposition~\ref{prop:invariance}) -- a property of the representation, not the attack: raw item-set Jaccard, no embedding, detects the same profiles at AUROC 0.92--0.97 where LightGCN collapses (\S\ref{sec:invariance}). That blind spot is not reliably bounded by checking any single dataset, architecture, or detection method; and the right response is therefore a framework that reports calibrated confidence in its own detections and is checked against exactly those axes. Building and testing VEIL, structured after Bou et al.'s Alignment Auditor~\cite{Bou2026}, produced the three findings stated in \S\ref{sec:intro} and reported in \S\ref{sec:results} -- Component~B's confidence $\alpha$ is non-monotonic in attack difficulty; donor-mimicry collapses detection from AUROC 0.936--1.000 to exact chance (0.48--0.51) through a traceable, already-known mechanism; and the severity of that collapse ranks differently by dataset, architecture, and detection method, with no axis predicting the others -- plus a validation check: Component~C's posterior contracts monotonically with evidence, run as a label-routed auditing experiment, not a deployable detector.

The practical upshot is that a single AUROC number, reported without auditing whether confidence degrades correctly under harder evidence or generalizes beyond the dataset, architecture, or detection method it was measured with, cannot distinguish a robust detector from one attack-design choice, dataset switch, or architecture swap away from chance. \S\ref{sec:limitations} bounds how far that claim travels.

\bibliographystyle{unsrtnat}
\bibliography{references}

\newpage
\appendix

\section{Component C, Unsupervised Routing}\label{app:unsup}

\paragraph{Unsupervised routing: a stable partition, but only weakly tied to the attackers.} We tested replacing Component~C's ground-truth routing with a stochastic variational-EM procedure on CiaoDVD's classic-attack (easy) configuration, the smallest diagnostic before any larger commitment. No true label was used anywhere in the fitting procedure; true labels were used only afterward, to evaluate how well the recovered assignment matched reality -- standard practice for validating an unsupervised method.

This is an important contrast to state plainly. Proposition~\ref{prop:invariance} and the mechanism in \S\ref{sec:invariance} describe information that is structurally absent from this encoder's representation: LightGCN never sees rating values in the first place, so no downstream method, supervised or not, can recover from $\phi$ what was never represented in it -- a limit of this representation, not of donor-mimicry attacks generally (a raw, embedding-free score does recover it; Appendix~\ref{app:labelfree}). The CiaoDVD classic-attack configuration is different: the organic-versus-attacker distinction is present in the representation and is not hard to recover with supervision -- labeled Component~C achieves AUROC 0.827--0.869 on this exact configuration. The open question here is narrower and different in kind: whether an unsupervised estimator, with no label access at all, can reliably find that same distinction on its own.

The answer is a qualified yes, with a precise limitation. Four label-free initializations were tried: $k$-means clustering, isolation-forest anomaly scoring, a genuinely random 50/50 split, and a principal-component threshold. All numbers below are for this one configuration (2,095 profiles, 273 injected). The original runs did not save their assignments, so we regenerated the seed-42 runs with identical code and settings, reproducing every recorded mixing weight $\hat\pi$, AUROC, and final ELBO to the printed precision.

\paragraph{k-means and PCA-threshold converge to the same partition with the roles swapped.} The two runs' hard assignments agree on 98.4\% of profiles once the organic and attacker labels are swapped (Pearson $r=-0.97$ between their posteriors; adjusted Rand index 0.94 between the partitions), a split of about 41\%/59\%. The PCA-threshold run started from a 16.7\% split and abandoned it (ARI 0.015 with its own converged partition). Their apparent contrast, AUROC 0.61 against 0.38 (post-hoc, true labels), is this swap: flipping the PCA run's roles gives 0.62. The two branches are near-exchangeable (they differ only through initialization and a weak minority prior), so the final ELBO is essentially the same for both orientations ($-164.86$ vs.\ $-164.00$, a gap smaller than the 6.6-nat spread across k-means restarts alone) and cannot choose between them. This is ordinary label switching in a two-component mixture, not two competing estimators of which one is right.

\paragraph{A label-free orientation rule fixes this.} Calling the smaller cluster ``attacker'' (equivalently $\hat\pi<0.5$, consistent with the model's generic minority prior) gives post-hoc AUROC 0.610--0.618 across all eight runs (each of the two initializations at seed 42 and three further seeds). The rule uses no labels, but we adopted it after seeing the swap, not in advance. Restarts are stable: pairwise ARI between runs is 0.94--0.98 both within each initialization and across the two, so the AUROC contrast is not run-to-run instability.

\paragraph{But the partition is only weakly tied to the attackers.} It agrees with the true labels at ARI 0.049. Its smaller side (41\% of profiles, against a true injection rate of 13\%) is 20.6\% injected against 7.8\% on the other side, roughly a $1.6\times$ enrichment, and that enrichment is what produces the AUROC. In the embedding the model was fit on, the injected profiles are not a cluster (silhouette $-0.06$ for the true partition, against 0.41 for this one), so a cluster-quality criterion prefers the wrong partition, and neither the ELBO, silhouette, nor restarts can reveal that the stable solution is not the attack split.

\paragraph{The other two initializations.} Random 50/50 ends at a genuinely different and weak solution (ARI 0.13 with the shared partition, silhouette 0.09, final ELBO $-1309$). Isolation-forest initialization ends at a related but distinct partition (ARI 0.60 with the shared one, same orientation, AUROC 0.529). The isolation-forest and local-outlier-factor detectors themselves flag 15.5\% and 12.7\% of profiles, and none of the 273 injected profiles is flagged by either: density-based anomaly detection finds real minority structure that is unrelated to the attack.

We read this as a limit on unsupervised recovery on this task, not an implementation gap. The fit is reproducible and its orientation is fixable by convention, but the partition it finds is only weakly related to the attackers, so unsupervised Component~C is a partial capability, not a replacement for label-routing. This is a one-configuration diagnostic and we make no claim that it generalizes.

\paragraph{A residual signal that a linear reward cannot express.} On ML-1M's full-ladder donor-mimicry configuration, a radial statistic on the frozen embeddings (distance from the embedding centroid) recovers a small residual signal, AUROC 0.58 averaged over VEIL's eight resampled test splits (0.59--0.60 for the closely related Isolation Forest and $k$-means-distance scores), against 0.49 for VEIL's Component~A on the same splits. The evidence is weak: only $k$-means-distance survives a Bonferroni correction over the 20 tests examined (the radial statistic alone, tested separately at $p=0.01$, does not), and on the easier matched (no-AUSH) variant the gap shrinks to a near-tie (0.56 vs.\ 0.53). VEIL's reward is linear in the frozen embedding and cannot express a radial displacement, and a label-trained linear probe on the same embeddings also stays at chance (0.47), so the residual is consistent with a function-class limit rather than a supervision gap; it is not evidence that a generic baseline outperforms VEIL.

\section{Label-Free Detectability of the Injected Profiles, Three Datasets}\label{app:labelfree}

This diagnostic asks whether the information VEIL uses on the easy attacks is also recoverable \emph{without labels} from the same frozen embeddings, and whether the answer depends on the dataset. It is not a competing detector: no score is tuned, and AUROC is reported \emph{raw} (higher score = more anomalous = predicted attacker), so a value below 0.5 means the injected profiles look \emph{more} normal than organic ones. We use four scores on the frozen LightGCN embeddings: mean distance to the 10 nearest neighbours (kNN), Local Outlier Factor with 20 neighbours (LOF), distance to the nearest of two $k$-means centres, and a radial statistic (distance from the embedding centroid). Isolation Forest, run on ML-1M only, tracked the $k$-means-distance and radial scores closely and is omitted. We add a fifth score that uses \emph{no embedding at all}: each profile's maximum Jaccard similarity, over its raw rated-item set, to any other profile's raw item set (self excluded) -- the obvious raw-data check, since a profile that nearly duplicates an existing user is an easy anomaly signal. Computed via sparse binary user-item matrix products, row-chunked (500 rows/chunk) so the full $N\times N$ similarity matrix is never materialized densely (Musical Instruments has $N=18{,}370$ profiles). For each of VEIL's 8 sweep seeds we recompute the scores on that seed's own re-split, re-standardized embeddings and evaluate on that seed's test split, so every label-free number is paired with VEIL's on the same split; Jaccard does not depend on an embedding or a split, so it is computed once per config and re-evaluated on each seed's test indices. Table~\ref{tab:labelfree} gives the 8-split means and Table~\ref{tab:bootstrap} full-pool bootstrap intervals. CiaoDVD and Musical Instruments have only the classic random+average+bandwagon config and its donor-mimicry variant (no AUSH config was run there), so the per-type results of Table~\ref{tab:pertype} use those three attack types; ML-1M additionally has an AUSH config.

\begin{table}[H]
\centering
\footnotesize
\setlength{\tabcolsep}{4pt}
\begin{tabular}{llcccccc}
\toprule
dataset & config & VEIL A/B/C & kNN & LOF & $k$-means & radial & Jaccard\\
\midrule
ML-1M & classic mix & 1.000/1.000/1.000 & 0.613 & 0.344 & 0.618 & 0.670 & 0.340\\
 & AUSH & 1.000/1.000/1.000 & 0.000 & 0.350 & 0.971 & 0.937 & 1.000\\
 & donor, full ladder & 0.494/0.488/0.481 & 0.478 & 0.547 & 0.603 & 0.577 & \textbf{0.974}\\
 & donor, matched (no AUSH) & 0.533/0.536/0.530 & 0.444 & 0.530 & 0.558 & 0.557 & \textbf{0.924}\\
\midrule
CiaoDVD & classic mix & 0.877/0.878/0.879 & 0.299 & 0.102 & 0.503 & 0.365 & 0.389\\
 & donor (no AUSH) & 0.846/0.848/0.846 & 0.477 & 0.472 & 0.582 & 0.631 & \textbf{0.942}\\
\midrule
MusInst & classic mix & 0.990/0.991/0.991 & 0.140 & 0.075 & 0.118 & 0.107 & 0.341\\
 & donor (no AUSH) & 0.805/0.808/0.808 & 0.360 & 0.410 & 0.500 & 0.504 & \textbf{0.940}\\
\bottomrule
\end{tabular}
\caption{Label-free anomaly scores vs.\ labeled VEIL, mixed-attack configs. All values are AUROC against the true labels, \emph{raw} (higher score = more anomalous = predicted attacker; never flipped, so a value below 0.5 means the injected profiles look \emph{more} normal than organic ones). VEIL: mean over its 8 sweep seeds (each on its own train/test re-split). kNN/LOF/$k$-means/radial: recomputed on each seed's own re-split and re-standardized embeddings and averaged over the same 8 test splits. Jaccard: raw item sets, no embedding, evaluated on the identical 8 test splits (\textbf{bold}: the four donor-mimicry configs, where Jaccard is the only score to detect what every embedding-based score -- VEIL included -- misses; Table~\ref{tab:donor}). No AUSH config exists for CiaoDVD or Musical Instruments (MusInst). Full-pool bootstrap intervals: Table~\ref{tab:bootstrap}.}
\label{tab:labelfree}
\end{table}

\begin{table}[H]
\centering
\scriptsize
\setlength{\tabcolsep}{3pt}
\begin{tabular}{llcccc}
\toprule
dataset & config & kNN & LOF & $k$-means & radial\\
\midrule
ML-1M & classic mix & 0.620 [0.569, 0.672] & 0.335 [0.304, 0.364] & 0.617 [0.580, 0.653] & 0.668 [0.626, 0.706]\\
 & AUSH & 0.000 [0.000, 0.000] & 0.355 [0.302, 0.411] & 0.972 [0.966, 0.979] & 0.939 [0.929, 0.949]\\
 & donor, full ladder & 0.471 [0.412, 0.540] & 0.536 [0.470, 0.604] & 0.597 [0.544, 0.652] & 0.580 [0.521, 0.636]\\
 & donor, matched (no AUSH) & 0.438 [0.400, 0.476] & 0.525 [0.491, 0.562] & 0.538 [0.504, 0.576] & 0.541 [0.503, 0.577]\\
\midrule
CiaoDVD & classic mix & 0.304 [0.275, 0.334] & 0.105 [0.090, 0.121] & 0.509 [0.487, 0.532] & 0.362 [0.323, 0.400]\\
 & donor (no AUSH) & 0.482 [0.447, 0.516] & 0.473 [0.436, 0.509] & 0.568 [0.534, 0.600] & 0.621 [0.590, 0.649]\\
\midrule
MusInst & classic mix & 0.142 [0.136, 0.148] & 0.073 [0.066, 0.080] & 0.119 [0.114, 0.124] & 0.108 [0.103, 0.114]\\
 & donor (no AUSH) & 0.365 [0.354, 0.376] & 0.415 [0.404, 0.427] & 0.503 [0.492, 0.516] & 0.508 [0.496, 0.519]\\
\bottomrule
\end{tabular}
\caption{Companion to Table~\ref{tab:labelfree}: label-free AUROC over the full pool of profiles (embeddings as cached, not re-split), with 95\% bootstrap intervals over profiles. Pool values differ slightly from the 8-split means because they use a single standardization and all profiles.}
\label{tab:bootstrap}
\end{table}

\begin{table}[H]
\centering
\footnotesize
\begin{tabular}{lllccccc}
\toprule
dataset & config & attack & VEIL A$^\dagger$ & kNN & LOF & $k$-means & radial\\
\midrule
ML-1M & classic mix & random & 0.999 & 0.928 & 0.364 & 0.807 & 0.896\\
 & & average & 0.999 & 0.930 & 0.392 & 0.828 & 0.906\\
 & & bandwagon & 1.000 & 0.000 & 0.279 & 0.232 & 0.221\\
 & AUSH & aush & 1.000$^\ddagger$ & 0.000 & 0.350 & 0.971 & 0.937\\
\midrule
CiaoDVD & classic mix & random & 0.825 & 0.447 & 0.098 & 0.538 & 0.141\\
 & & average & 0.825 & 0.456 & 0.114 & 0.505 & 0.207\\
 & & bandwagon & 0.966 & 0.002 & 0.094 & 0.466 & 0.737\\
\midrule
MusInst & classic mix & random & 0.987 & 0.212 & 0.087 & 0.146 & 0.147\\
 & & average & 0.984 & 0.211 & 0.103 & 0.144 & 0.149\\
 & & bandwagon & 1.000 & 0.000 & 0.037 & 0.066 & 0.027\\
\bottomrule
\end{tabular}
\caption{Per-attack-type AUROC on the classic configs (positives: that attack type only; negatives: organic profiles), raw as in Table~\ref{tab:labelfree}. Label-free columns are 8-split means. $^\dagger$VEIL Component~A per type is from the saved posterior on a \emph{single} split (seed 42), reproducing the recorded overall AUROC; it is only approximately comparable to the 8-split label-free means. $^\ddagger$AUSH is a single attack type, so this VEIL entry is the 8-seed mean.}
\label{tab:pertype}
\end{table}

\paragraph{(a) Some score separates every attack type on every dataset; what varies is how many scores do.} Taking each type's best score with the orientation supplied by the labels (a descriptive quantity, not a detector), separability is 0.93 / 0.93 / 1.00 (random / average / bandwagon) on ML-1M's classic config, 0.90 / 0.89 / 1.00 on CiaoDVD, and 0.91 / 0.90 / 1.00 on Musical Instruments. On Musical Instruments all four scores separate every type (at least 0.78). On CiaoDVD only LOF does (0.89--0.91 on every type): $k$-means distance is at chance (0.47--0.54 raw) and kNN separates only bandwagon (0.998; 0.54--0.55 on the other two).

\paragraph{(b) Orientation ambiguity replicates and strengthens: the sign is dataset-dependent, not only attack-dependent.} On ML-1M's classic config, random and average are conventionally oriented for three of the four scores (kNN 0.93, radial 0.90--0.91, $k$-means 0.81--0.83), bandwagon is inverted for every score, and AUSH is inverted for kNN and LOF. On Musical Instruments every score on every type is inverted (all twelve type-by-score AUROCs are at most 0.21, and the mixed config gives 0.075--0.140), so a conventionally oriented label-free detector would perform far \emph{worse} than chance there. CiaoDVD is inverted in nine of twelve cells (LOF 0.09--0.11 on every type). The sign that would be needed cannot be read off the data without labels.

\paragraph{(c) The mixing collapse is conditional on sign conflicts, not a general property of mixing.} On ML-1M's classic config, kNN, $k$-means distance, and radial have opposite signs across types (radial: 0.90, 0.91, 0.22), and their mixed AUROC is 0.61--0.67, far below the per-type values. CiaoDVD's radial score shows the same conflict (0.14 and 0.21 against 0.74; mixed 0.365). Where the signs agree, mixing does not collapse the score: LOF on ML-1M (per-type 0.28--0.39, mixed 0.34), LOF on CiaoDVD (per-type 0.09--0.11, mixed 0.10, i.e.\ 0.90 once oriented), and every score on Musical Instruments (mixed, oriented: 0.86--0.93). CiaoDVD's kNN is not a sign conflict: all three types are below 0.5, and its mixed 0.299 reflects two near-chance types (0.45, 0.46) diluting one perfectly inverted type (0.002).

\paragraph{The raw-profile Jaccard baseline detects donor-mimicry where every embedding-based score, including VEIL, does not.} Item-set Jaccard needs no embedding: it is computed directly on each profile's raw rated-item set. On all four donor-mimicry configs across all three datasets it reaches raw AUROC 0.924--0.974 (8-seed mean; Table~\ref{tab:labelfree}), on the identical profiles where VEIL and every embedding-based label-free score sit at or near chance. Average precision is more moderate (0.50--0.57 raw) because Jaccard similarity is symmetric: a donor-mimicry profile's nearest neighbour is its clone, but the true donor's nearest neighbour is that same clone right back, so every true donor scores as ``anomalous'' as the profile it was copied into. Table~\ref{tab:donor} isolates this: flagging organic profiles whose best match is an injected profile at Jaccard $\ge0.5$ (the same threshold Appendix~\ref{app:mf} uses to identify true donors) and excluding them from the negative class (a diagnostic that uses the labels to identify donors, not a detector) raises AUROC to 0.9997--1.0000 and AP to 0.994--1.000 on every config. The number of organic profiles flagged this way is large relative to the injected count -- from 90 flagged (ML-1M full ladder, 91 injected) to 1,773 (Musical Instruments, 2,397 injected) -- confirming the donor-symmetry effect is real and substantial, not a rounding artifact, and that the raw (not donor-excluded) number is the fair one to headline: a deployed detector does not know in advance which organic users are donors.

\begin{table}[H]
\centering
\small
\begin{tabular}{lrcc}
\toprule
config & organic flagged/total & raw AUROC / AP & donor-excl.\ AUROC / AP\\
\midrule
ML-1M, full ladder & 90 / 1,812 & 0.974 / 0.515 & 1.0000 / 1.000\\
ML-1M, matched (no AUSH) & 257 / 1,812 & 0.924 / 0.503 & 1.0000 / 1.000\\
CiaoDVD, donor & 227 / 1,822 & 0.942 / 0.569 & 0.9999 / 0.999\\
Musical Instruments, donor & 1,773 / 15,973 & 0.940 / 0.549 & 0.9997 / 0.994\\
\bottomrule
\end{tabular}
\caption{Donor-effect diagnostic: an organic profile is flagged as a likely true donor if its best raw-Jaccard match is an injected profile at similarity $\ge0.5$. ``raw'' includes all organic profiles in the negative class (the number we headline); ``donor-excl.'' drops the flagged organic profiles from the negative class, isolating how much the symmetry effect costs. Both columns are 8-seed means, same test splits as Table~\ref{tab:labelfree}.}
\label{tab:donor}
\end{table}

On the classic (non-donor) configs Jaccard is not uniformly useful: raw AUROC is near or below chance (0.340--0.389) on plain classic mixes and strongly inverted on AUSH+heterogenize (0.044), but perfect on un-heterogenized AUSH (1.000) -- config-dependent, like every other label-free score in this appendix. The score has no tunable parameter.

\paragraph{The ML-1M donor-mimicry result does not generalize -- for the other four label-free scores.} Restricting to kNN/LOF/$k$-means/radial, the best score is at or slightly above VEIL's own near-chance value on ML-1M (0.603 vs.\ 0.494 on the full ladder; 0.558 vs.\ 0.533 on the matched variant; significance is discussed in Appendix~\ref{app:unsup}). On CiaoDVD and Musical Instruments VEIL does not collapse on the donor configs (0.846 and 0.805) and among those four scores outperforms every one of them: the best is 0.631 on CiaoDVD (radial; $k$-means distance 0.582, kNN and LOF below 0.5) and 0.504 on Musical Instruments (the other three at or below chance). Jaccard is the exception to all of this, and the more important comparison for what a real attacker would try first.

\paragraph{A label-assisted score edging VEIL is not evidence that a baseline beats it.} On CiaoDVD's classic config, LOF with the orientation supplied by the labels reaches 0.895 on the full pool (95\% interval $[0.879, 0.910]$; 0.898 averaged over the 8 splits), against VEIL's 0.877 (Component~A; 0.878 and 0.879 for B and C). The interval reaches VEIL's value at its lower edge, so this is at most a slight edge, and it uses label knowledge (the sign) that VEIL does not need. It is consistent with the function-class theme of Appendix~\ref{app:unsup}: a density-based score is not expressible as a linear reward, and CiaoDVD's classic config is where VEIL's separability is weakest among the easy configs.

\paragraph{Caveats.} The per-type VEIL numbers in Table~\ref{tab:pertype} come from a saved posterior on a single split (seed 42) and are only approximately comparable to the 8-split label-free means. Oriented values use labels to choose the sign and describe available information only. All results use LightGCN embeddings with $k=10$ neighbours (LOF: 20), fixed in advance; in a neighbourhood-size sweep (not shown) density scores change strongly with scale (for example, kNN on ML-1M's AUSH config is 0.000 for every $k$ up to 100 and 0.96 at $k=400$), so both the strength and the sign of density scores depend on the scale chosen.

\section{Full Calibration Table (A/B/C, All Seed-Swept Configs)}\label{app:calib}

8-seed mean AUROC/ECE (A, B) and AUROC/ECE (C); Brier and NLL given for Component~A only here for space -- full Brier/NLL/reliability-bin data for every component is in the supplementary per-config sweep files (\texttt{sweep\_vi\_seed.json}), not retyped here. 95\% CIs (Student-$t$, $n-1=7$ df) are in those files, not reproduced in this table for space; Table~\ref{tab:calib} in the main text gives Component~A's CIs for the same configs.

\begin{table}[H]
\centering
\footnotesize
\setlength{\tabcolsep}{4pt}
\begin{tabular}{lcccccccc}
\toprule
config & A AUROC & A ECE & A Brier & A NLL & B AUROC & B ECE & C AUROC & C ECE\\
\midrule
\cfg{supgnn\_base} & 0.997 & 0.006 & 0.006 & 0.056 & 0.999 & 0.002 & 0.999 & 0.002\\
\cfg{lightgcn\_base} & 1.000 & 0.006 & 0.005 & 0.047 & 1.000 & 0.006 & 1.000 & 0.005\\
\cfg{lightgcn\_aush} & 1.000 & 0.000 & 0.000 & 0.000 & 1.000 & 0.000 & 1.000 & 0.000\\
\cfg{lightgcn\_aush\_het} & 0.945 & 0.090 & 0.071 & 0.287 & 0.946 & 0.087 & 0.945 & 0.077\\
\cfg{lightgcn\_aush\_het\_donor} & 0.494 & 0.374 & 0.232 & 0.679 & 0.488 & 0.375 & 0.481 & 0.366\\
Musical Instruments (easy) & 0.990 & 0.050 & 0.038 & 0.139 & 0.991 & 0.049 & 0.991 & 0.048\\
Musical Instruments (hard) & 0.805 & 0.263 & 0.183 & 0.544 & 0.808 & 0.262 & 0.808 & 0.261\\
CiaoDVD (easy) & 0.877 & 0.177 & 0.137 & 0.472 & 0.878 & 0.177 & 0.879 & 0.175\\
CiaoDVD (hard) & 0.846 & 0.187 & 0.151 & 0.494 & 0.848 & 0.187 & 0.846 & 0.186\\
ML-1M matched (no AUSH) & 0.533 & 0.326 & 0.246 & 0.690 & 0.536 & 0.326 & 0.530 & 0.324\\
\bottomrule
\end{tabular}
\caption{Calibration across every seed-swept config (all now $\ge$8 seeds).}
\label{tab:calibfull}
\end{table}

\section{Component A Full Per-Config Table}\label{app:compA}

Appendix tables round to 4 decimal places (finer than the main text's 3 significant figures); full-\texttt{float} precision for every cell is in the supplementary results file.

\begin{table}[H]
\centering
\footnotesize
\setlength{\tabcolsep}{3.5pt}
\begin{tabular}{lccccc|cccc}
\toprule
 & \multicolumn{5}{c|}{test} & \multicolumn{4}{c}{train}\\
config & AUROC & AP & Acc & F1 & pairs-Brier/ECE & AUROC & AP & Acc & F1\\
\midrule
\cfg{supgnn\_base} & 0.9982 & 0.9614 & 0.9984 & 0.9939 & 0.0029 / 0.0176 & 0.9987 & 0.9872 & 0.9959 & 0.9843\\
\cfg{lightgcn\_base} & 0.9993 & 0.9948 & 0.9920 & 0.9697 & 0.0001 / 0.0004 & 1.0000 & 1.0000 & 0.9986 & 0.9948\\
\cfg{lightgcn\_aush} & 1.0000 & 1.0000 & 1.0000 & 1.0000 & 0.0000 / 0.0000 & 1.0000 & 1.0000 & 1.0000 & 1.0000\\
\cfg{lightgcn\_aush\_het} & 0.9424 & 0.4234 & 0.9492 & 0.5246 & 0.0428 / 0.0318 & 0.9781 & 0.6188 & 0.9655 & 0.6406\\
\cfg{lightgcn\_aush\_het\_donor} & 0.4996 & 0.0511 & 0.9142 & 0.0755 & 0.3554 / 0.2697 & 0.7433 & 0.1067 & 0.9189 & 0.1563\\
\bottomrule
\end{tabular}
\caption{Component~A, full per-config detection metrics (test and train splits) at 4-decimal precision, plus held-out preference-pair Brier/ECE.}
\label{tab:compA}
\end{table}

\section{Component B Full Per-Config Table}\label{app:compB}

\begin{table}[H]
\centering
\footnotesize
\setlength{\tabcolsep}{3.5pt}
\begin{tabular}{lccc|ccccc}
\toprule
config & $\alpha$ & $\lambda_{\min}$ & $\lambda_{\max}$ & test AUROC & test AP & test Acc & test F1 & train AUROC\\
\midrule
\cfg{supgnn\_base} & 3.2168 & 0.0268 & 0.4751 & 0.9982 & 0.9512 & 0.9984 & 0.9939 & 1.0000\\
\cfg{lightgcn\_base} & 0.4209 & 0.0729 & 0.3904 & 0.9996 & 0.9973 & 0.9920 & 0.9697 & 1.0000\\
\cfg{lightgcn\_aush} & 0.4025 & 0.1317 & 0.2386 & 1.0000 & 1.0000 & 1.0000 & 1.0000 & 1.0000\\
\cfg{lightgcn\_aush\_het} & 0.7044 & 0.0512 & 0.4866 & 0.9442 & 0.4391 & 0.9527 & 0.5574 & 0.9801\\
\cfg{lightgcn\_aush\_het\_donor} & 0.5412 & 0.1076 & 0.4984 & 0.5134 & 0.0523 & 0.9177 & 0.0784 & 0.7382\\
\bottomrule
\end{tabular}
\caption{Component~B, full per-config learned $\alpha$, ARD precision range ($\lambda_{\min}$, $\lambda_{\max}$), and detection metrics at 4-decimal precision (seed-42 headline run).}
\label{tab:compB}
\end{table}

\begin{table}[H]
\centering
\small
\begin{tabular}{lcccc}
\toprule
config & $\alpha$ mean & std & SE & 95\% CI ($t$, df$=7$)\\
\midrule
\cfg{lightgcn\_base} & 0.3861 & 0.0174 & 0.0062 & [0.3715, 0.4006]\\
\cfg{lightgcn\_aush} & 0.3286 & 0.0604 & 0.0213 & [0.2781, 0.3791]\\
\cfg{lightgcn\_aush\_het} & 0.6877 & 0.0610 & 0.0216 & [0.6367, 0.7387]\\
\cfg{lightgcn\_aush\_het\_donor} & 0.5514 & 0.0850 & 0.0301 & [0.4803, 0.6225]\\
\bottomrule
\end{tabular}
\caption{Component~B learned $\alpha$, 8-seed aggregation (seeds 42, 1--7; sample std, SE $=$ std$/\sqrt{8}$). Seeds vary pair sampling, split, and VI initialization over one trained Stage-0 feature cache per config; encoder-retraining variance is not captured. Source: supplementary results file.}
\label{tab:alpha8}
\end{table}

\section{Component C Full Per-Config Table + Evidence Ablation}\label{app:compC}

\begin{table}[H]
\centering
\footnotesize
\setlength{\tabcolsep}{3.5pt}
\begin{tabular}{lcc|ccccc}
\toprule
config & $\alpha_{\mathrm{org}}$ & $\alpha_{\mathrm{att}}$ & test AUROC & test AP & test Brier & test ECE & train AUROC\\
\midrule
\cfg{supgnn\_base} & 3.2373 & 3.0016 & 0.9982 & 0.9512 & 0.0016 & 0.0017 & 1.0000\\
\cfg{lightgcn\_base} & 0.4670 & 0.2973 & 0.9997 & 0.9978 & 0.0063 & 0.0062 & 1.0000\\
\cfg{lightgcn\_aush} & 0.6916 & 0.3056 & 1.0000 & 1.0000 & 0.0000 & 0.0000 & 1.0000\\
\cfg{lightgcn\_aush\_het} & 0.8285 & 0.4898 & 0.9364 & 0.4642 & 0.0707 & 0.0837 & 0.9848\\
\cfg{lightgcn\_aush\_het\_donor} & 0.5398 & 0.3934 & 0.5075 & 0.0631 & 0.2384 & 0.3889 & 0.7465\\
\bottomrule
\end{tabular}
\caption{Component~C, full per-config organic/attacker branch $\alpha$ and combined-detection metrics at 4-decimal precision.}
\label{tab:compC}
\end{table}

Component~C attacker-branch evidence-accumulation ablation (\S\ref{sec:contraction}), full precision, base LightGCN config. The ablation script does not record which feature cache it read; we identify it as the base-LightGCN cache because the ablation's injected-count ceiling (191) matches that config's injected count exactly, so this identification is an inference, not a logged fact (Table~\ref{tab:ablation}).

\begin{table}[H]
\centering
\small
\begin{tabular}{cccccc}
\toprule
$n_{\text{injected}}$ & $n_{\text{train pairs}}$ & $\log|\Sigma|$ & $\bar\sigma$ & $\alpha$ & branch AUROC (test)\\
\midrule
20 & 300 & 78.4868 & 1.8578 & 0.1831 & 0.9990\\
50 & 750 & 59.0742 & 1.5937 & 0.2108 & 0.9992\\
100 & 1500 & 37.6224 & 1.3496 & 0.2527 & 0.9998\\
150 & 2250 & 20.2082 & 1.1777 & 0.2697 & 0.9997\\
191 & 2865 & 15.9541 & 1.1388 & 0.2973 & 0.9997\\
\bottomrule
\end{tabular}
\caption{Component~C attacker-branch posterior contraction vs.\ injected evidence count, full precision.}
\label{tab:ablation}
\end{table}

\section{Matrix Factorization (\texorpdfstring{$n_{\text{layers}}=0$}{n\_layers=0}) -- Full Results, and the Donor-Distance Check}\label{app:mf}

\paragraph{Method.} \texttt{lightgcn\_layers=0} on the existing \texttt{LightGCN}/\texttt{\_lightgcn\_embeddings} path -- no separate module. Confirmed by full training-loop inspection to be an exact reduction to classical BPR matrix factorization~\cite{Rendle2009}: the propagation loop never executes, so the returned embedding is the raw layer-0 embedding table; negative sampling and the BPR loss are layer-count-independent by construction; $L_2$ regularization was already computed on raw layer-0 embeddings, not something disabled for this comparison.

\begin{table}[H]
\centering
\footnotesize
\setlength{\tabcolsep}{3.5pt}
\begin{tabularx}{\textwidth}{lcccc>{\raggedright\arraybackslash}X}
\toprule
config & Stage 0 (s) & A AUROC & B AUROC & C AUROC & LightGCN comparison\\
\midrule
\cfg{ml1m\_mf\_base} & 191.0 & 0.9941 & 0.9944 & 0.9956 & indistinguishable (both near-ceiling)\\
\cfg{ml1m\_mf\_aush} & 221.2 & 1.0000 & 1.0000 & 1.0000 & both saturate\\
\cfg{ml1m\_mf\_aush\_het} & 220.4 & 0.9451 & 0.9465 & 0.9547 & nearly identical\\
\cfg{ml1m\_mf\_aush\_het\_donor}$^\ddagger$ & 1100.6 & 0.6320 & 0.6424 & 0.6305 & MF substantially more robust (LightGCN: 0.481--0.494)\\
\cfg{ciaodvd\_mf\_classic} & 23.6 & 0.8137 & 0.8119 & 0.8148 & LightGCN notably better (0.877--0.879)\\
\cfg{ciaodvd\_mf\_het\_donor} & 19.3 & 0.7644 & 0.7675 & 0.7614 & LightGCN notably better (0.846--0.848)\\
\bottomrule
\end{tabularx}
\caption{MF full ladder, all 6 configs. $^\ddagger$Raw wall-clock includes an anomalous $\approx890$\,s machine-level freeze (confirmed by a simultaneously-silent independent resource monitor, so not workload-specific); freeze-adjusted compute $\approx210$\,s, consistent with the other MF configs. Reported as measured, not excluded.}
\label{tab:mf}
\end{table}

Architecture ranking is dataset/config-dependent, not uniform: MF is notably more robust than LightGCN specifically on ML-1M's donor-mimicry point, but consistently worse than LightGCN on both CiaoDVD points -- plausibly MF underfitting CiaoDVD's much sparser per-profile signal ($\approx15$ ratings/user vs.\ ML-1M's $\approx165$).

\paragraph{Donor-distance check, full protocol (\S\ref{sec:mechanism}).} On \cfg{ml1m\_aush\_het\_donor}, we reconstructed the poisoned dataset (deterministic given seed=42; identical for both the LightGCN and MF cached runs) and recovered each injected profile's true donor via item-set Jaccard overlap -- donor-mimicry copies the donor's item set almost wholesale, so the true donor's overlap with an injected profile is far higher than any other organic user's by construction (mean Jaccard 0.8517, median 0.9600; 81 of 91 injected profiles confidently matched at Jaccard $\ge0.5$, 10 excluded as ambiguous). We then measured injected-to-donor distance in each encoder's frozen \texttt{phi}, plus an injected-to-random-organic-user control.

\begin{table}[H]
\centering
\small
\begin{tabular}{lcc}
\toprule
metric (mean $\pm$ SEM, $n=81$) & LightGCN & MF\\
\midrule
injected$\to$\textbf{donor} Euclidean dist. & $1.9826\pm0.0361$ & $3.0931\pm0.0628$\\
injected$\to$\textbf{donor} cosine similarity & $0.9673\pm0.0020$ & $0.9203\pm0.0037$\\
injected$\to$\textbf{random-organic} Euclidean dist.\ (control) & $10.9134\pm0.3751$ & $11.0304\pm0.2411$\\
injected$\to$\textbf{random-organic} cosine similarity (control) & $0.0633\pm0.0434$ & $0.0133\pm0.0198$\\
\bottomrule
\end{tabular}
\caption{Injected-to-donor vs.\ injected-to-random-organic embedding distance, LightGCN vs.\ MF, \cfg{ml1m\_aush\_het\_donor}. Paired difference (MF $-$ LightGCN Euclidean distance to true donor): $1.1104\pm0.0433$ SEM, $n=81$, approx.\ paired $t=25.65$.}
\label{tab:donordist}
\end{table}

Both encoders place injected profiles far closer to their true donor than to a random organic user, confirming the donor match is meaningful and not an artifact of the matching procedure; the control distances are nearly identical between encoders (10.91 vs.\ 11.03), which rules out a general embedding-scale difference as the explanation for the donor-gap result. The donor-specific gap (LightGCN 1.98 vs.\ MF 3.09) is real, tight (SEM $\le0.07$ on both sides), and in the direction the propagation-smoothing hypothesis predicts. This is correlational and single-config, not a controlled ablation isolating propagation depth from every other difference between the two fits -- reported in \S\ref{sec:mechanism} as supporting evidence for the mechanistic account, not as independent proof of it.

\paragraph{Sequential encoders vs.\ MF, full analysis (\S\ref{sec:arch}).} On Musical Instruments' order-preserving donor-mimicry config, Table~\ref{tab:arch} ranks MF $>$ SASRec $>$ mean-pool $>$ LightGCN on every component. LightGCN's position is explained by the same propagation mechanism as the donor-distance result above: it is the only one of the four architectures that mixes an injected profile's embedding with its neighbors', including its donor's, and the other three -- which compute a representation directly from a profile's own interactions, with no graph diffusion -- all score well above it. The ordering among those three is not explained by that mechanism, or by any other we tested: MF has no sequence information at all and a free per-user parameter vector, the opposite design from both sequence encoders, yet scores highest of the four, ahead of SASRec (95\% CI does not overlap on any component, e.g.\ Component~C $[0.938, 0.943]$ vs.\ $[0.900, 0.904]$), which in turn beats the order-agnostic mean-pool control. We report this secondary ordering as measured, not explained, and flag it as an open question rather than assert a cause.

\section{Classical Baseline (A2): Full Results}\label{app:a2}

\paragraph{Method.} RDMA/\allowbreak WDMA/\allowbreak LengthVar hand-crafted rating-value features~\cite{Chirita2005} + logistic regression (\texttt{class\_weight="balanced"}, $C{=}1.0$), fit on the \emph{same} poisoned dataset and stratified train/test split Components A/B/C use for each config -- same label access at train time, same held-out test set, same metric set -- deliberately excluding DegSim-style neighbor-similarity features (an $O(N^2)$ scaling risk flagged and excluded from this baseline). Operates on raw rating triples, never on the frozen embedding \texttt{phi}; Stage~0 (GNN training) is never invoked.

\begin{table}[H]
\centering
\footnotesize
\setlength{\tabcolsep}{3.5pt}
\begin{tabularx}{\textwidth}{llccc>{\raggedright\arraybackslash}X}
\toprule
dataset & point & A2 AUROC & A2 AP & VEIL AUROC range & note\\
\midrule
CiaoDVD & easy (classic) & 0.8397 & 0.3451 & 0.877--0.879 & comparable, VEIL slightly ahead\\
CiaoDVD & hard (het+donor) & 0.6171 & 0.1630 & 0.846--0.848 & A2 collapses far more; VEIL more robust here\\
ML-1M & easy (base) & 0.9822 & 0.8085 & 0.999--1.000 & comparable, both near-ceiling\\
ML-1M & hard (matched, no AUSH) & 0.5838 & 0.1669 & 0.530--0.536 & both collapse substantially\\
\bottomrule
\end{tabularx}
\caption{A2 (RDMA/\allowbreak WDMA/\allowbreak LengthVar + logistic regression) vs.\ VEIL, all 4 tested points. $n_{\text{test}}=629$ (CiaoDVD) / 626 (ML-1M), $n_{\text{injected,test}}=82$ throughout -- split sizes match VEIL's own exactly, confirming the fair-comparison design.}
\label{tab:a2}
\end{table}

Not a uniform result: on ML-1M the classical baseline also collapses substantially under donor-mimicry, supporting this paper's structural (not VEIL-specific) framing; on CiaoDVD the classical baseline collapses \emph{more} than VEIL does, the reverse asymmetry from what either simple hypothesis (baseline is robust vs.\ baseline also collapses) would have predicted alone.

\section{Wall-Clock Summary}\label{app:wallclock}

\begin{table}[H]
\centering
\footnotesize
\setlength{\tabcolsep}{4pt}
\begin{tabular}{lccccc}
\toprule
config & Stage 0 & Comp.\ A & Comp.\ B & Comp.\ C & total\\
\midrule
\cfg{supgnn\_base} & 2697.5s & 8.4s & 4.2s & 8.2s & 2719.6s (Stage0 log) / 20.9s (ABC log)\\
\cfg{lightgcn\_base} & 2661.6s & 7.7s & 4.7s & -- & 2679.3s (Stage0) / 12.4s (AB log)\\
\cfg{lightgcn\_aush} & 2570.9s & 5.4s & 4.7s & 10.5s & 2591.6s\\
\cfg{lightgcn\_aush\_het} & 2272.8s & 4.8s & 4.2s & 8.4s & 2290.2s\\
\cfg{lightgcn\_aush\_het\_donor} & 5239.2s & 4.6s & 4.4s & 9.5s & 5257.7s\\
\bottomrule
\end{tabular}
\caption{Per-stage wall-clock. Hardware: i3-1115G4 (2 cores / 4 threads), CPU-only, \texttt{threads=4}.}
\label{tab:wallclock}
\end{table}

\paragraph{Note on the \cfg{lightgcn\_base} row.} No single log covers Stage~0 plus Components A, B, and C for this config. Component~C was run for it separately and its wall-clock was not captured, hence the blank entry and the split totals.

\section{Hyperparameters}\label{app:hyper}

The default configuration file, verbatim:

{\footnotesize
\begin{Verbatim}
run:     { seed: 42, threads: 4, device: cpu }
data:    { dataset: ml-1m, users_frac: 0.30, injection_rate: 0.05,
           filler_budget_pct: 0.03, max_filler_items: 120, n_target_items: 5,
           test_frac: 0.30, aush_pretrain_epochs: 12, aush_hidden_dim: 128,
           aush_lambda_push: 1.0, het_pop_exponent: 0.6, het_body_quantile: 0.85,
           het_target_rating_lo: 4.0, het_target_rating_hi: 5.0,
           het_filler_noise: 0.6, het_pref_boost: 5.0, het_donor_min_ratings: 10 }
gnn:     { embedding: user_fused, hidden_dim: 64, epochs: 20, lr: 0.001,
           lightgcn_dim: 64, lightgcn_layers: 3, lightgcn_epochs: 30,
           lightgcn_lr: 0.001, lightgcn_batch_size: 2048, lightgcn_reg: 0.0001 }
pairs:   { pairs_per_injected: 20, pairs_per_anchor: 15, matching: random }
vi:      { steps: 3000, batch_size: 256, lr: 0.01, mc_samples: 4, sigma0: 1.0,
           alpha: 1.0 }                                         # Component A
hier:    { steps: 2000, lr: 0.005, mc_samples: 4, tau1: 1.0, tau2: 1.0,
           beta: 1.0, gamma1: 1.0, gamma2: 1.0, alpha_init: 1.0 } # Component B
branch:  { steps: 2000, lr: 0.005, mc_samples: 4,
           eval_mc_samples: 32 }                                # Component C
\end{Verbatim}
}

\paragraph{Dataset sizes after 5-core filtering.} ML-1M: 999,611 interactions (30\% of users sampled before injection). Amazon Musical Instruments: 53,246 users / 27,637 items / 472,422 reviews (30\% of users sampled). CiaoDVD: 1,822 users / 2,069 items / 28,374 reviews, not subsampled, since its 5-core survivor is already smaller than ML-1M's 30\% subsample.

\paragraph{Attack ladder, per-config detail} (Table~\ref{tab:ladder} has the results). Five configs of increasing difficulty:
\begin{enumerate}[itemsep=1pt]
\item \cfg{supgnn\_base} / \cfg{lightgcn\_base} -- random + average + bandwagon @ 5\% injection each.
\item \cfg{lightgcn\_aush} -- AUSH (GAN-based, \texttt{src/attacks/aush.py}), 12 pretrain epochs (\texttt{data.aush\_pretrain\_epochs: 12}).
\item \cfg{lightgcn\_aush\_het} -- AUSH + \texttt{HeterogenizeAttack} wrapper: per-profile size from the organic degree distribution, stochastic popularity/preference-weighted filler, jittered 4.0--5.0 target ratings, per-profile (bias, noise) rating style (\texttt{het\_per\_profile\_style: true}).
\item \cfg{lightgcn\_aush\_het\_donor} -- adds \texttt{--donor-mimicry}: each fake profile copies a real organic user's full (items, ratings) and splices in a random 1--3-item target subset (\texttt{het\_donor\_min\_ratings: 10}).
\item An 8-seed sweep (42, 1, 2, \ldots, 7) repeating config 4 to quantify seed variance at the collapse point (\texttt{sweep\_vi\_seed.py}).
\end{enumerate}

\paragraph{Why LightGCN is blind to rating values, line-by-line} (moved here from \S\ref{sec:mechanism} for space): \texttt{src/models/lightgcn.py}'s \texttt{build\_norm\_adj} (line 40) builds the bipartite user-item adjacency from raw (\texttt{user\_idx}, \texttt{item\_idx}) pairs, explicitly deduplicating (\texttt{np.unique(np.stack([row, col], axis=1), axis=0)}, line 71) into one unweighted edge per (user, item) pair -- rating \emph{value} is never read. \texttt{bpr\_loss} (line 153) takes \texttt{users, pos\_items, neg\_items} -- item \textbf{IDs} only, no rating tensor in the signature at all.

\section{Algorithm Boxes}\label{app:algo}

Pseudocode below mirrors the prose in \S\ref{sec:veil} and the actual control flow of the Component A/B/C implementations and the shared VI module; function names are the real ones in the codebase, not renamed for presentation.

\paragraph{Component A} (\texttt{component\_a.py:run}):

{\footnotesize
\begin{Verbatim}[commandchars=\\\{\}]
input: cached features Phi (organic + injected), config cfg
pairs_train, pairs_test <- make_ab_pairs(feats, cfg, split)
                                        # Delta-phi = phi(x+) - phi(x-)
q <- ELBOModel(dim, alpha=1.0 FIXED, sigma0=cfg.vi.sigma0)
                                        # mu, log-sigma^2 parameters
for step in 1..cfg.vi.steps:                                  # 3000
    batch <- sample_minibatch(pairs_train, cfg.vi.batch_size) # 256
    theta_samples <- reparameterize(q.mu, q.sigma, mc_samples)
                                        # cfg.vi.mc_samples=4
    L <- mean_s[ sum_i log sigmoid(alpha * theta_samples_s^T Delta-phi_i) ]
         - KL(q || N(0, sigma0^2 I))
    Adam.step(maximize L)                                     # lr=cfg.vi.lr=0.01
report <- \{posterior: q, pairs_test metrics, detection_\{train,test\} metrics\}
save_report(report); save(q, "component_a_posterior.pt")
\end{Verbatim}
}

\paragraph{Component B} (\texttt{component\_b.py:build\_model}, \texttt{:run}):

{\footnotesize
\begin{Verbatim}[commandchars=\\\{\}]
input: same features, warm-start <- load("component_a_posterior.pt")
q <- ELBOModel(dim, alpha ~ Gamma(tau1, tau2), lambda_d ~ Gamma(gamma1, gamma2),
               mu_d ~ N(0, (beta * lambda_d)^-1), warm_start=A's (mu, sigma))
for step in 1..cfg.hier.steps:                                # 2000
    batch <- sample_minibatch(pairs_train, ...)
    (theta_s, alpha_s, lambda_s) <- reparameterize(q, mc_samples)
    L <- mean_s[ sum_i log sigmoid(alpha_s * theta_s^T Delta-phi_i) ]
         - KL(q(theta,alpha,lambda) || hierarchical prior)     # extended ELBO
    Adam.step(maximize L)                                     # lr=cfg.hier.lr=0.005
report <- \{posterior (incl. learned alpha, lambda_min/max), pairs/detection metrics\}
save_report(report); save(q, "component_b_posterior.pt")
\end{Verbatim}
}

\paragraph{Component C} (\texttt{component\_c.py:\_branch\_pools}, \texttt{:\_fit\_branch}, \texttt{:\_combined\_metrics}, \texttt{:run}):

{\footnotesize
\begin{Verbatim}[commandchars=\\\{\}]
input: same features, ground-truth organic/injected labels
organic_pool, attacker_pool <- _branch_pools(feats, split)    # routed by label
for anchor in \{"organic", "attacker"\}:
    branch[anchor] <- _fit_branch(cfg, feats, anchor)
        # same as Component B; cfg.branch.steps=2000; warm-started from A
r_org(x) <- branch["organic"].reward_projection(x)            # theta_org^T phi(x)
r_att(x) <- branch["attacker"].reward_projection(x)           # theta_att^T phi(x)
score(x) <- r_att(x) - r_org(x)
(w, b) <- _logistic_fit(score[train], y[train])               # 1-D logistic calibration
P(injected | x) <- sigmoid(w * score(x) + b)
report <- _combined_metrics(cfg, feats, branch, split)        # AUROC/AP/Brier/ECE
save_report(report)
\end{Verbatim}
}

\end{document}